\documentclass[10pt,twocolumn,letterpaper]{article}

\usepackage{iccv}
\usepackage{microtype}
\usepackage{times}
\usepackage{epsfig}
\usepackage{graphicx}
\usepackage{amsmath}
\usepackage{amssymb}
\usepackage{caption}
\usepackage{subcaption}
\usepackage{booktabs}
\usepackage{makecell}
\usepackage{authblk}

\usepackage[pagebackref=true,breaklinks=true,letterpaper=true,colorlinks,linkcolor=blue,citecolor=blue,bookmarks=false]{hyperref}

\iccvfinalcopy 

\def\iccvPaperID{1234}

\ificcvfinal\fi
\begin{document}

\title{Guided Proofreading of Automatic Segmentations for Connectomics}

\author[1,2]{Daniel Haehn\thanks{Corresponding author,  \url{haehn@seas.harvard.edu}}}
\author[1,2]{Verena Kaynig}
\author[3]{James Tompkin}
\author[2]{Jeff W. Lichtman}
\author[1,2]{Hanspeter Pfister}
\affil[1]{Harvard Paulson School of Engineering and Applied Sciences, Cambridge, MA 02138, USA}
\affil[2]{Harvard Brain Science Center, Cambridge, MA 02138, USA}
\affil[3]{Brown University, Providence, RI 02912, USA}

\maketitle

\begin{abstract}
Automatic cell image segmentation methods in connectomics produce merge and
split errors, which require correction through proofreading. Previous research
has identified the visual search for these errors as the bottleneck in
interactive proofreading. To aid error correction, we develop two classifiers
that automatically recommend candidate merges and splits to the user. These
classifiers use a convolutional neural network (CNN) that has been trained with
errors in automatic segmentations against expert-labeled ground truth. Our
classifiers detect potentially-erroneous regions by considering a large context
region around a segmentation boundary. Corrections can then be performed by a
user with yes/no decisions, which reduces variation of information $7.5\times$ faster than previous
proofreading methods. We also present a fully-automatic mode that uses a
probability threshold to make merge/split decisions. Extensive experiments using
the automatic approach and comparing performance of novice and expert users
demonstrate that our method performs favorably against state-of-the-art
proofreading methods on different connectomics datasets.
\end{abstract}

\section{Introduction}

In connectomics, neuroscientists annotate neurons and their connectivity within
3D volumes to gain insight into the functional structure of the brain. Rapid
progress in automatic sample preparation and electron microscopy (EM)
acquisition techniques has made it possible to image large volumes of brain
tissue at nanometer resolution. With a voxel size of
$4\times4\times40~\text{nm}^3$, a cubic millimeter volume is one petabyte of
data. With so much data, manual annotation is not feasible, and automatic
annotation methods are needed~\cite{jain2010,Liu2014,GALA2014,kaynig2015large}.

\begin{figure}[t]
\begin{center}
  \includegraphics[width=\linewidth]{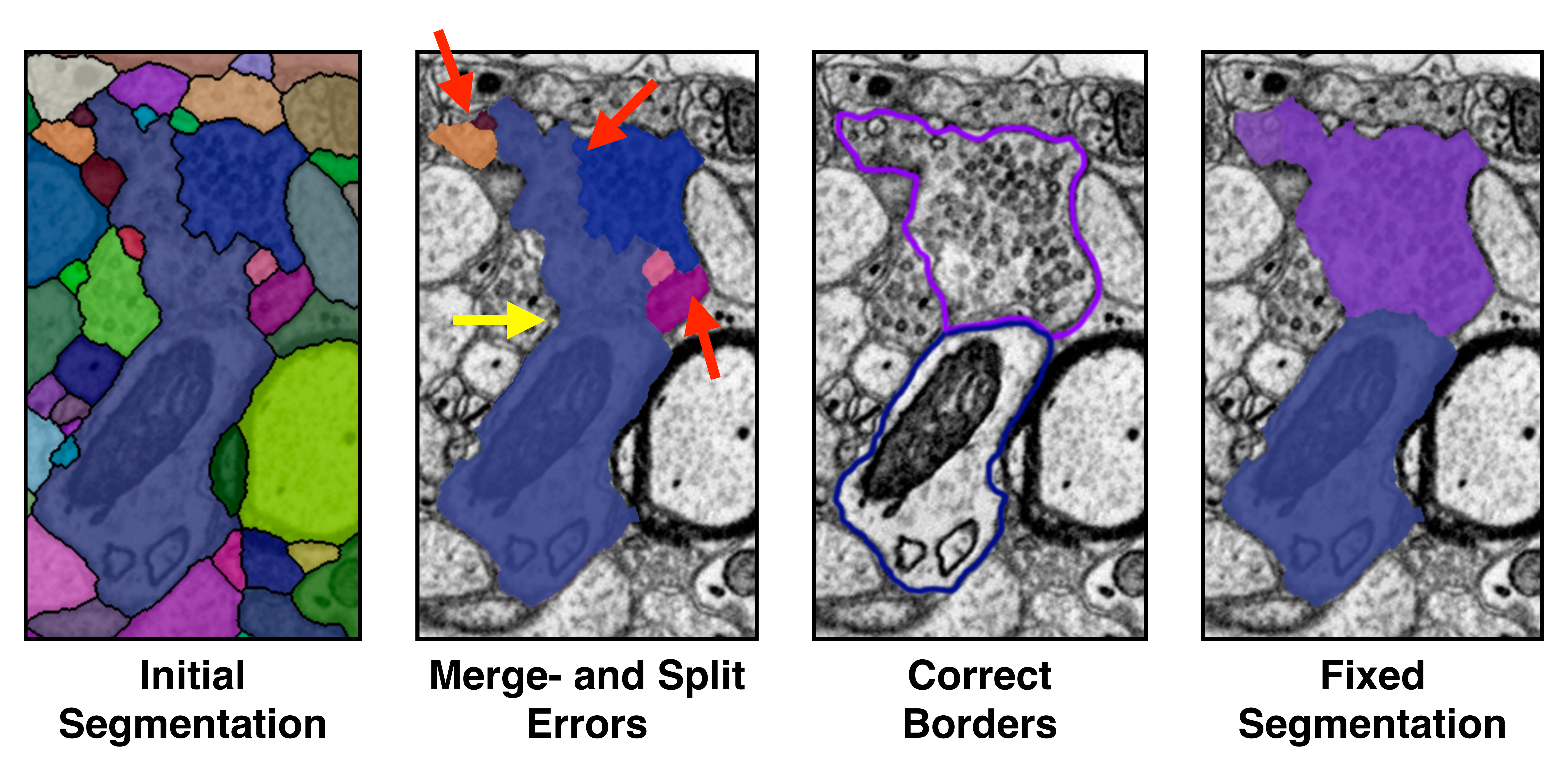}
\end{center}
\vspace{-4mm}
   \caption{The most common proofreading corrections are fixing split errors (red arrows) and merge errors (yellow arrow). A fixed segmentation matches the cell borders.}
\label{fig:merge_and_slit_errors}
\end{figure}

Automatic annotation by segmentation and classification of brain tissue is
challenging~\cite{isbi_challenge} and all available methods make errors, so 
the results must be \emph{proofread} by humans. This crucial
task serves two purposes: 1) to correct errors in the segmentation, and 2) to
increase the body of labeled data from which to train better automatic
segmentation methods. Recent proofreading tools provide intuitive user
interfaces to browse segmentation data in 2D and 3D and to identify and manually
correct errors~\cite{markus_proofreading,raveler,mojo2,haehn_dojo_2014}. Many
kinds of errors exist, such as inaccurate boundaries, but the most common are
\emph{split errors}, where a single segment is labeled as two, and \emph{merge
errors}, where two segments are labeled as one
(Fig.~\ref{fig:merge_and_slit_errors}). With user interaction, split errors can
be joined, and the missing boundary in a merge error can be defined with
manually-seeded watersheds~\cite{haehn_dojo_2014}. However, the visual
inspection to find errors takes the majority of the time, even with
semi-automatic correction tools~\cite{proofreading_bottleneck}.

Our goal is to automatically detect potential split and merge errors to reduce visual
inspection time. Further, to reduce correction time, we propose
corrections to the user to accept or reject. We call this process \textit{guided
proofreading}.

We train a classifier for split error detection with a convolutional neural network
(CNN). This takes as input patches of membrane segmentation probabilities, cell
segmentation masks, and boundary masks, and outputs a split-probability score. As we
must process large data, this classifier only operates on cell boundaries, which
reduces computation over methods that analyze every pixel. For merge errors, we
invert and reuse the split classification network, and ask it to rate a
set of generated boundaries that hypothesize a split. 

Possible erroneous regions are sorted by their score, and a candidate correction is generated for each
region. Then, a user works through this list of regions and corrections. In a
forced choice setting, the user either selects a correction or skips it to
advance to the next region. In an automatic setting, errors with a high probability are automatically corrected first, given an appropriate
probability threshold, after which the user would take over. Finally, to test
the limits of performance, we create an oracle which only accepts corrections
that improve the segmentation, based on knowledge of the ground truth. This is
guided proofreading with a perfect user.

We evaluate these methods on multiple connectomics datasets. For the forced
choice setting, we perform a quantitative user study with 20 novice users who
have no previous experience of proofreading EM data. We ask participants to
proofread a small segmentation volume in a fixed time frame. In a
between-subjects design, we compare guided proofreading to the semi-automatic
\textit{focused proofreading} approach by Plaza~\cite{focused_proofreading}. In
addition, we compare against the manual interactive proofreading tool
\textit{Dojo} by Haehn~\etal~\cite{haehn_dojo_2014}. We also asked four domain
experts to use guided proofreading and focused proofreading for comparison.

This paper makes the following contributions.
First, we present a CNN-based boundary classifier for split errors, plus a merge
error classifier that inverts the split error classifier. This is used to
propose merge error corrections, removing the need to manually draw the missing
edge. These classifiers perform well without much training data, which is
expensive to collect for connectomics data.
Second, we developed a guided proofreading approach to correcting segmentation
volumes, and an assessment scenario comparing forced-choice interaction with
automatic and oracle proofreading.
Third, we present the results of a quantitative user study assessing
guided proofreading. Our method is able to reduce segmentation
error faster than state-of-the-art semi-automatic tools for both novice and
expert users.

Guided proofreading is applicable to all existing automatic segmentation methods that
produce a label map. As such, we believe that our approach is a promising
direction to proofread segmentations more efficiently and better tackle large
volumes of connectomics imagery.

\begin{figure*}[t]
\centering
\includegraphics[width=\linewidth]{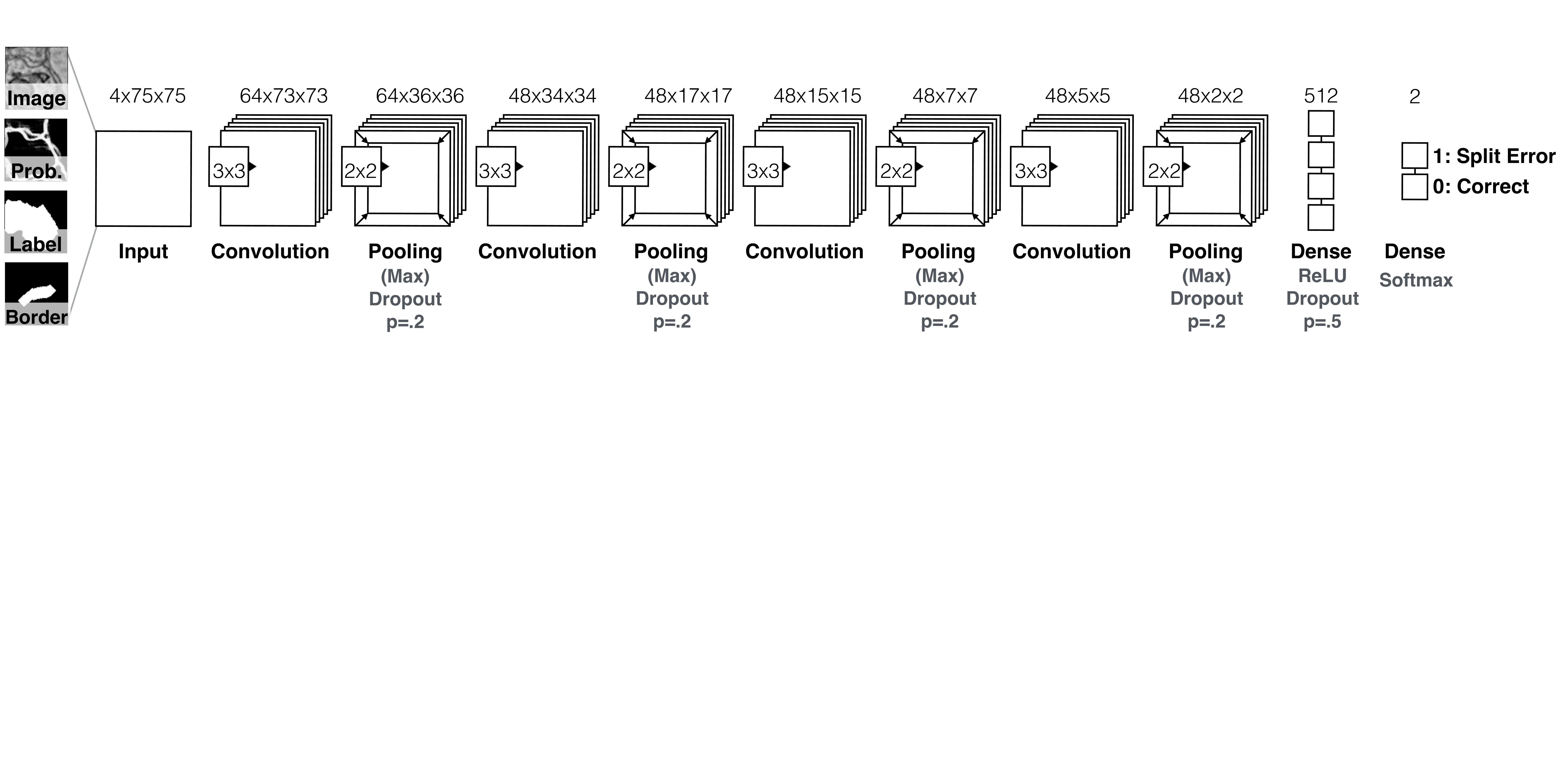}
\caption{Our guided proofreading classifiers use a traditional CNN architecture of four convolutional layers, each followed by max pooling and dropout regularization. The 4-channel input patches are labelled either as correct splits or as split errors.}
\label{fig:architecture}
\end{figure*}

\section{Related Work}

\textbf{Automatic Segmentation.} Multi-terabyte EM brain volumes require automatic segmentation~\cite{jain2010,Liu2014,NunezIglesias2013Machine,GALA2014}, but can be hard to classify due to ambiguous intercellular space: the 2013 IEEE ISBI neurites 3D segmentation challenge~\cite{isbi_challenge} showed that existing algorithms that learn from expert-segmented training data still exhibit high error rates.

Many works tackle this problem. NeuroProof~\cite{neuroproof2013} decreases error rates by learning an agglomeration on over-segmentations of images based on a random forest classifier. Vazquez-Reina \etal~\cite{amelio_segmentation} consider whole EM volumes rather than a per-section approach, then solve a fusion problem with a global context. Kaynig \etal~\cite{kaynig10} propose a random forest classifier coupled with an anisotropic smoothing prior in a conditional random field framework with 3D segment fusion. Bogovic \etal~\cite{BogovicHJ13} learn 3D features unsupervised, and show that they can be better than by-hand designs.

It is also possible to learn segmentation classification features directly from images with CNNs. Ronneberger \etal~\cite{RonnebergerFB15} use a contracting/expanding CNN path architecture to enable precise boundary localization with small amounts of training data. Lee \etal~\cite{lee2015recursive} recursively train very deep networks with 2D and 3D filters to detect boundaries.
All these approaches make good progress; however, in general, proofreading is still required to correct errors.

\paragraph{Interactive Proofreading.} While proofreading is very time consuming, it is fairly easy for humans to perform corrections through splitting and merging segments. One expert tool is Raveler, introduced by Chklovskii~\etal~\cite{chklovskii2010, raveler}. Raveler is used today by professional proofreaders, and it offers many parameters for tweaking the process. Similar systems exist as products or plugins to visualization systems,~\eg, V3D~\cite{proofreading_bottleneck} and AVIZO~\cite{markus_proofreading}.

Recent papers have tackled the problem of proofreading massive datasets through crowdsourcing with novices~\cite{saalfeld09,anderson2011,Giuly2013DP2}. One popular platform is EyeWire, by Kim \etal~\cite{eyewire_nature}, where participants earn virtual rewards for merging over-segmented labelings to reconstruct retina cells.

Between expert systems and online games sit Mojo and Dojo, by Haehn
\etal~\cite{haehn_dojo_2014,Neuroblocks}, which use simple scribble interfaces
for error correction. Dojo extends this to distributed proofreading via a
minimalistic web-based user interface. The authors define requirements for
general proofreading tools, and then evaluate the accuracy and speed of Raveler,
Mojo, and Dojo through a quantitative user study (Sec.~3 and
4, ~\cite{haehn_dojo_2014}). Dojo had the highest performance. In this paper, we
use Dojo as a baseline for interactive proofreading.

All interactive proofreading solutions require the user to find potential errors manually, which takes the majority of the time~\cite{proofreading_bottleneck,haehn_dojo_2014}. Recent papers propose computer-aided proofreading systems to quicken this visual search task.

\paragraph{Computer-aided Proofreading.} Uzunbas \etal~\cite{uzunbas} showed that potential labeling errors can be found by considering the merge tree of an automatic segmentation method. The authors track uncertainty throughout the automatic labeling by training a conditional random field. This segmentation technique produces uncertainty estimates, which inform potential regions for proofreading to the user. While this applies to isotropic volumes, more work is needed to apply it to the typically anisotropic connectomics dataset volumes.

Karimov \etal~\cite{karimov_guided_volume_editing} propose guided volume editing, which measures the difference in histogram distributions in image data to find potential split and merge errors in the corresponding segmentation. This lets expert users correct labeled computer-tomography datasets, using several interactions per correction. To correct merge errors, the authors create a large number of superpixels within a single segment and then successively group them based on dissimilarities. We were inspired by this approach, but instead we generate single watershed boundaries to handle the intracellular variance in high-resolution EM images (Sec.~\ref{sec:methods}).

Most closely related to our approach is the work of
Plaza~\cite{focused_proofreading}, who proposed \textit{focused proofreading}.
This method generates affinity scores by analyzing a region adjacency graph
across slices, then finds the largest affinities based on a defined impact
score. This yields edges of potential split errors which can be presented to the
proofreader. Plaza reports that additional manual work is required to find and
correct merge errors. Focused proofreading builds upon
NeuroProof~\cite{neuroproof2013} as its agglomerator, and is open source with
integration into Raveler. As the closest related work, we wish to use this
method as a baseline to evaluate our approach (Sec.~\ref{sec:evaluation}).
However, we separate the backend affinity score calculation from the Raveler
expert-level front end, and present our own novice-friendly interface
(Sec.~\ref{sec:evaluation}).

\section{Method}
\label{sec:methods}

\subsection{Split Error Detection}
\label{sec:spliterrordetection}

We build a split error classifier with output $p$ using a CNN to check whether an edge within an existing automatic segmentation is valid ($p=0$) or not ($p=1$). Rather than analyzing every input pixel, the classifier operates only on segment boundaries, which requires less pixel context and is faster. In contrast to Bogovic \etal~\cite{BogovicHJ13}, we work with 2D slices rather than 3D volumes. This enables proofreading prior or in parallel to a computationally expensive stitching and 3D alignment of individual EM images.

\paragraph{CNN Architecture.} Boundary split error detection is a binary classification task since the boundary is either correct or erroneous. However, in reality, the score $p$ is between 0 and 1. In connectomics, classification complexity arises from hundreds of different cell types, rather than from the classification decision itself. Intuitively, this yields a wider architecture with more filters rather than a deeper architecture with more layers. We explored different architectures---including residual networks~\cite{resnet}---with brute force parameter searches and precision and recall comparisons (see supplementary materials). Our final CNN configuration for split error detection has four convolutional layers, each followed by max pooling with dropout regularization to prevent overfitting due to limited training data (Fig.~\ref{fig:architecture}).

\paragraph{Classifier Inputs.} To train the CNN, we consider boundary context in the decision making process via a $75\times75$ patch over the center of an existing boundary. This size covers approximately $80\%$ of all boundaries in the 6~nm Mouse S1 AC3 Open Connectome Project dataset. If the boundary is not fully covered, we sample up to 10 non-overlapping patches along the boundary, and average the resulting scores weighted by the boundary length coverage per patch.

\begin{figure}[t]
\centering
\includegraphics[width=\linewidth]{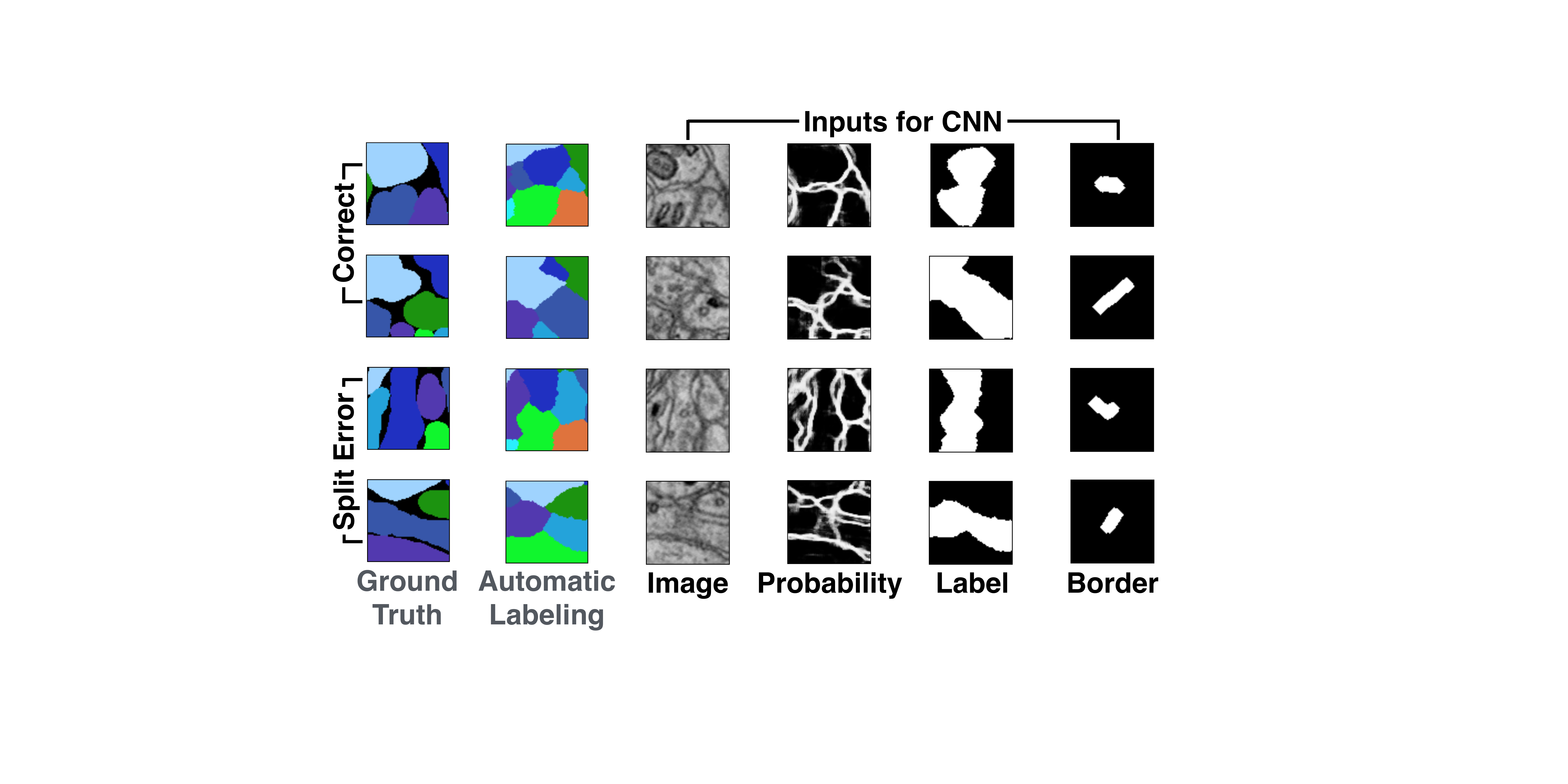}
\caption{Example inputs for learning correct splits and split errors (candidate segmentation versus the ground truth). Image, membrane probabilities, merged binary labels, and a dilated border mask provide 4-channel input patches.}
\label{fig:cnn_inputs}
\end{figure}

Similar to Bogovic~\etal~\cite{BogovicHJ13}, we use grayscale image data, corresponding boundary probabilities, and a single binary mask combining the two neighboring labels as inputs to our CNN. However, we observed that the boundary probability information generated from EM images is often misleading due to noise or artifacts in the data. This can result in merge errors within the automatic segmentation. To better direct our classifier to train on the true boundary, we extract the border between two segments. Then, we dilate this border by 5 pixels to consider slight edge ambiguities as well as cover extra-cellular space, and use this binary mask as an additional input. This creates a stacked 4-channel input patch. Fig.~\ref{fig:cnn_inputs} shows examples of correct and erroneous input patches and their corresponding automatic segmentation and ground truth.

\subsection{Merge Error Detection}

Identification and correction of merge errors is more challenging than finding and fixing split errors, because we must look inside segmentation regions for missing or incomplete boundaries and then propose the correct boundary. However, we can reuse the same trained CNN for this task. Similar to guided volume editing by Karimov~\etal~\cite{karimov_guided_volume_editing}, we generate potential borders within a segment. For each segmentation label, we dilate the label by 20 pixel and generate 50 potential boundaries through the region by randomly placing watershed seed points at opposite sides of the label boundary. We perform watershed on the inverted grayscale EM image. This yields 50 candidate splits.

\begin{figure}[t]
\centering
\includegraphics[width=\linewidth]{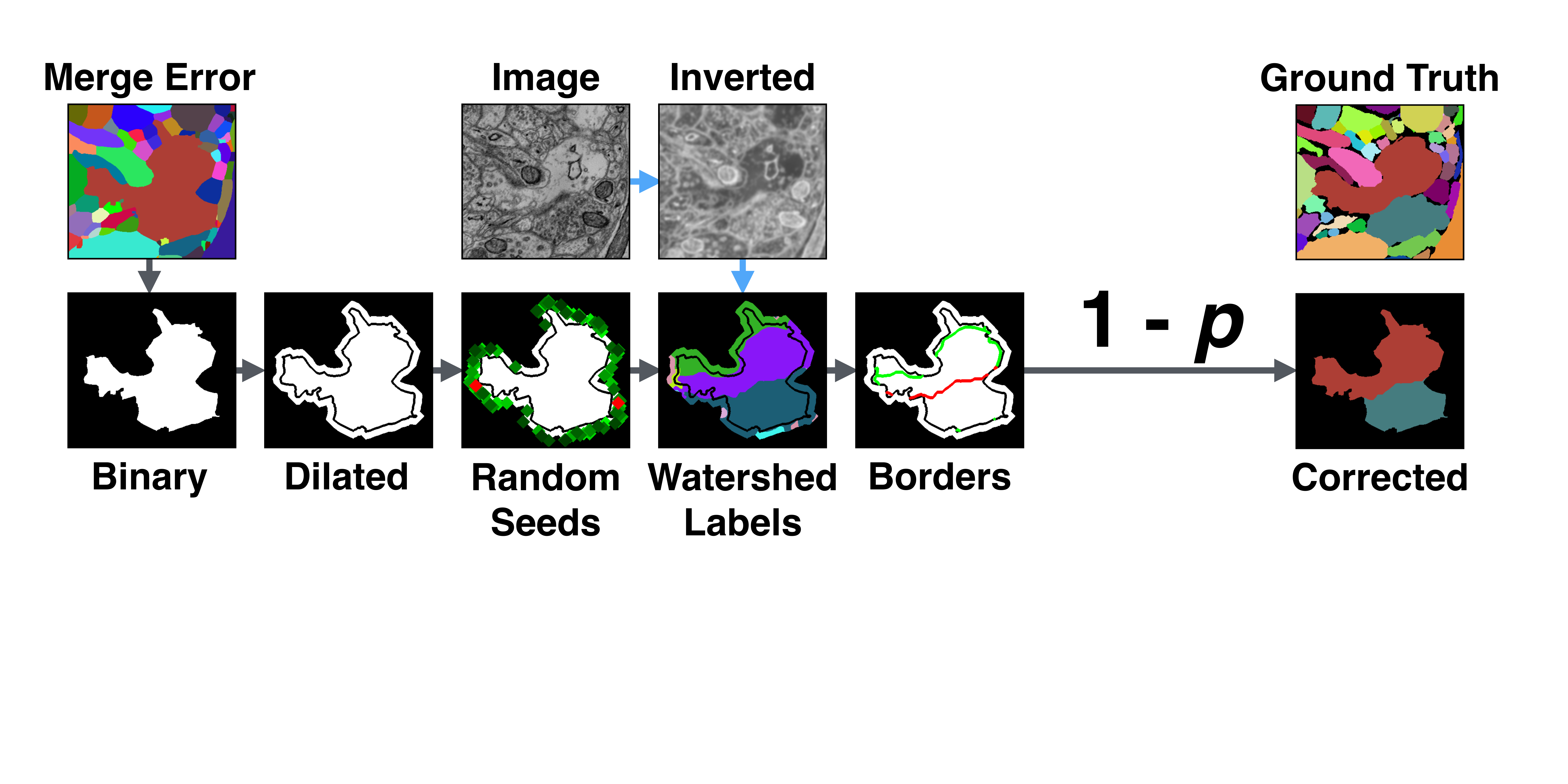}
\caption{Merge error detection: Potential borders are generated using inverted images by randomly placing watershed seeds (green) on the boundary of a dilated segment. The best ranked seeds and border (both in red) result in the shown error correction.}

\label{fig:merge_error}
\end{figure}

Dilation of the segment prior to watershed is motivated by our observation that the generated splits tend to attach to real membrane boundaries. These boundaries are then individually rated using our split error classifier. For this, we invert the probability score such that a correct split (previously encoded as $p=0$) is most likely a candidate for a merge error (now encoded as $p=1$). In other words, if a generated boundary is ranked as correct, it probably should be in the segmentation. Fig. \ref{fig:merge_error} illustrates this procedure. Pseudo code is available as supplemental material to promote understanding.

\subsection{Error Correction}
\label{sec:errorcorrection}

We combine the proposed classifiers to perform corrections of split and merge errors in automatic segmentations. For this, we first perform merge error detection for all existing segments in a dataset and store the inverted rankings $1-p$ as well as potential corrections. After that, we perform split error detection and store the ranking $p$ for all neighboring segments in the segmentation. Then, we sort the merge and split error rankings separately from highest to lowest. For error correction, first we loop through the potential merge error regions and then through the potential split error regions. During this process, each error is now subject to a yes/no decision which can be provided in different ways:

\paragraph{Selection oracle.} If ground truth data is available, the selection oracle \textit{knows} whether a possible correction improves an automatic segmentation. This is realized by simply comparing the outcome of a correction using a defined measure. The oracle only accepts corrections which improve the automatic segmentation---others get
discarded. This is guided proofreading with a perfect user, and allows us to assess the upper limit of improvements.

\paragraph{Automatic selection with threshold.} The decision whether to accept or reject a potential correction is taken by comparing rankings to a threshold $p_t$. If the inverted score $1-p$ of a merge error is higher than a threshold $1-p_t$, the correction is accepted. Similarly, a correction is accepted for a split error if the ranking $p$ is higher than $p_t$. Our experiments have shown that the threshold $p_t$ is the same for merge and split errors for a balanced classifier that has been trained on equal numbers of correct and error patches.

\paragraph{Forced choice setting.} We present a user with the choice to accept or reject a correction. All potential split errors are seen. Inspecting all merge errors is not possible for users due to the sheer amount of generated borders. Therefore, we only present merge errors that have a probability threshold higher than $1-p_t$.

\noindent \newline In all cases, a decision has to be made to advance to the next possible erroneous region. If a merge error correction was accepted, the newly found boundary is added to the segmentation data. This partially updates the merge error and split error ranking with respect to the new segment. If a split error correction was accepted, two segments are merged in the segmentation data and the disappearing segment is removed from all error rankings. Then, we perform merge error detection on the now larger segment and update the ranking. We also update the split error rankings to include all new neighbors, and re-sort. The error with the next highest ranking then forces a choice.

\subsection{User Interface}

\begin{figure}[t]
\includegraphics[width=\linewidth]{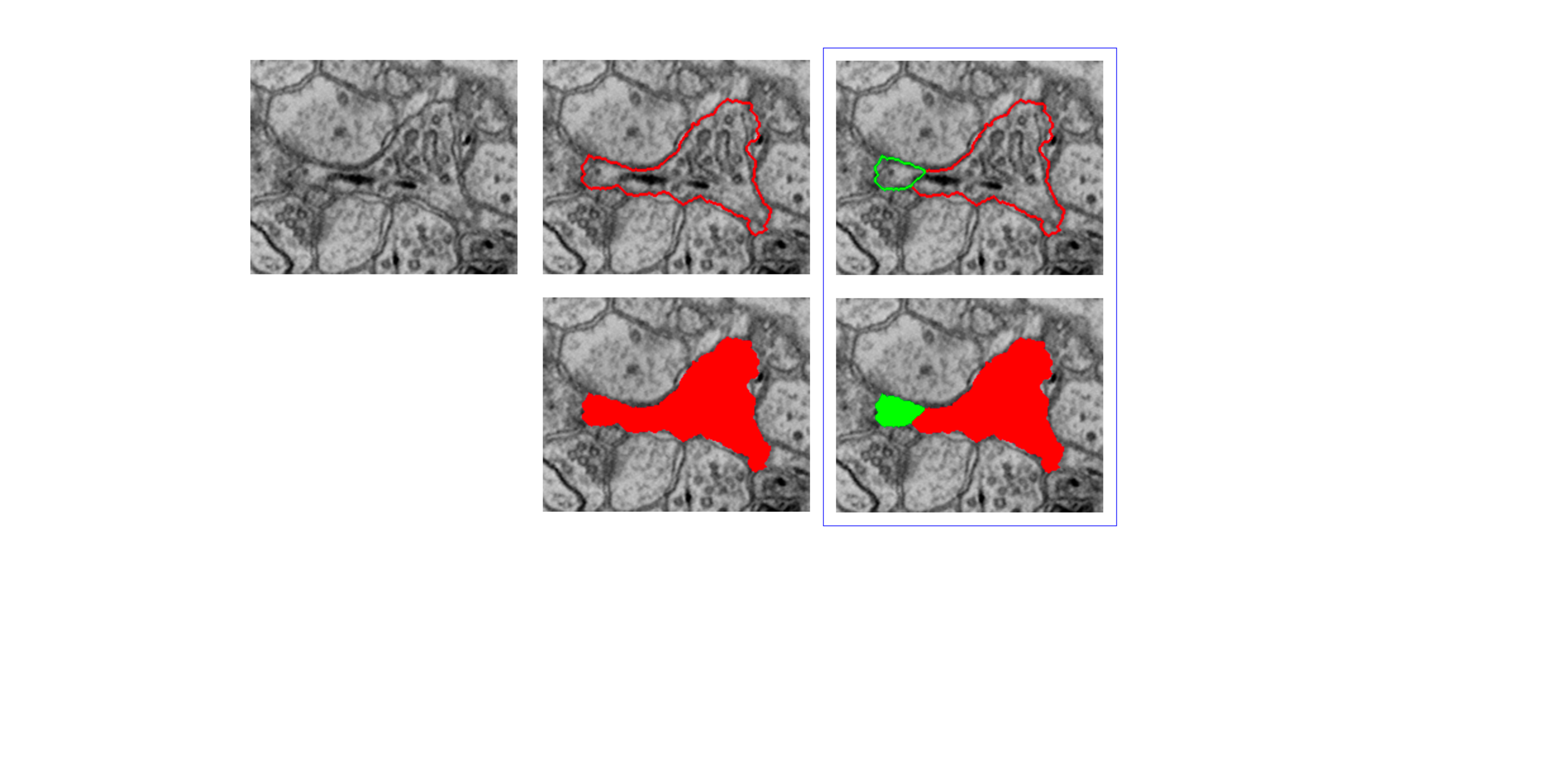}
\caption{User interface. A candidate error region is shown on the left. The user must choose between the region being a split error which needs correcting (center) or not (right). Confirming the choice advances to the next potential error.} 
\label{fig:ui}
\end{figure}

We integrate guided proofreading into an existing large data connectomics workflow. The web-based system is designed with a novice-friendly user interface (Fig.~\ref{fig:ui}). We show the current labeling of a cell boundary outline and its proposed correction overlayed on EM image data. The user cannot distinguish the current labeling from the proposed correction to avoid selection bias. We also show a solid overlay of the current and the proposed labeling. In addition, we show the image without overlays to provide an unoccluded view. User interaction is simple and involves one mouse click on either the current labeling or the correction. After interaction, the next potential error is shown.

\section{Evaluation}
\label{sec:evaluation}

We evaluate guided proofreading on multiple different real-world connectomics datasets of different species. All datasets were acquired using either serial section electron microscopy (ssEM) or serial section transmission electron microscopy (ssTEM). We perform experiments with the selection oracle, with automatic selection with threshold, and in the forced choice setting via a between-subjects user study with both novice and expert participants.

\subsection{Datasets}

\paragraph{L. Cylinder.} We use the left part of the 3-cylinder mouse cortex volume of Kasthuri \etal~\cite{kasthuri2015saturated} ($2048\times2048\times300$ voxels). The tissue is dense mammalian neuropil from layers 4 and 5 of the S1 primary somatosensory cortex, acquired using ssEM. The dataset resolution is $3\times3\times30~\text{nm}^3\text{/voxel}$. Image data and a manually-labeled expert `ground truth' segmentation is publicly available\footnote{\scriptsize{\url{https://software.rc.fas.harvard.edu/lichtman/vast/}}}.

\paragraph{AC4 subvolume.} This is part of a publicly-available dataset of mouse cortex that was published for the ISBI 2013 challenge ``SNEMI3D: 3D Segmentation of neurites in EM images''. The dataset resolution is $6\times6\times30~\text{nm}^3\text{/voxel}$ and it was acquired using ssEM. Haehn~\etal~\cite{haehn_dojo_2014} found the most representative subvolume ($400\times400\times10$ voxels) of this dataset with respect to the distribution of object sizes, and used it for their interactive connectomics proofreading tool experiments. We use their publicly available data, labeled ground truth, and study findings\footnote{\scriptsize{\url{http://rhoana.org/dojo/}}}.

\paragraph{Automatic segmentation pipeline.}
We use a state-of-the-art method to create a dense automatic segmentation of the data. Membrane probabilities are generated using a CNN based on the U-net architecture (trained exclusively on different data than the GP classifiers)~\cite{RonnebergerFB15}. The probabilities are used to seed watershed and generate an oversegmentation using superpixels. Agglomeration is then performed by the GALA active learning classifier with a fixed agglomeration threshold of 0.3~\cite{nunez2014graph}. We describe this approach in the supplemental material.

\subsection{Classifier Training}

We train our split error classifier on the L. Cylinder dataset. We use the first 250 sections of the data for training and validation. For n-fold cross validation, we select one quarter of this data and re-select after each epoch. We minimize cross-entropy loss and update using stochastic gradient descent with Nesterov momentum~\cite{nesterov}. To generate training data, we identify correct regions and split errors in the automatic segmentation by intersection with ground truth regions. This is required since extracellular space is not labeled in the ground truth, but is in our dense automatic segmentation. From these regions, we sample 112,760 correct and 112,760 split error patches with 4-channels (Sec.~\ref{sec:spliterrordetection}). The patches are normalized. To augment our training data, we rotate patches within each mini-batch by $k*90$ degrees with randomly chosen integer $k$. The training parameters such as filter size, number of filters, learning rate, and momentum are the result of intuition and experience, studying recent machine learning research, and a limited brute force parameter search (see supplementary material). 

Table~\ref{tab:parameters} lists the final parameters. Our CNN configuration results in 171,474 learnable parameters. We assume that training has converged if the validation loss does not decrease for 50 epochs. We test the CNN by generating a balanced set of 8,780 correct and 8,780 error patches using unseen data of the left cylinder dataset. 

\begin{table}[t]
\caption{Training parameters, cost, and results of our guided proofreading classifier versus focused proofreading by Plaza~\cite{focused_proofreading}. Both methods were trained on the same mouse brain dataset using the same hardware (Tesla X GPU).}
\small{
\begin{tabular}{ll}
	\toprule
	\begin{tabular}{l}
		\textbf{Guided Proofreading} \\ \midrule
		\emph{Parameters} \\ \midrule
		Filter size: 3x3 \\ No. Filters 1: 64 \\ No. Filters 2--4: 48 \\ Dense units: 512 \\ Learning rate: 0.03--0.00001\\ Momentum: 0.9--0.999\\Mini-Batchsize: 128 \\
	\end{tabular}
	&
	\begin{tabular}{l}
		\vspace{0.2mm} \\
		\midrule
		\emph{Results---Test Set} \\ \midrule Cost [m]: 383 \\ Val. loss: 0.0845 \\ Val. acc.: 0.969 \\ Test. acc.: 0.94 \\ Prec./Recall: 0.94/0.94 \\ F1 Score: 0.94 \\ ~ \\
	\end{tabular}
\end{tabular}

\vspace{0.5mm}
\begin{tabular}{ll}
	\toprule
	\begin{tabular}{l}
		\textbf{Focused Proofreading}\\ \midrule
		\emph{Parameters} \\ \midrule
		Iterations: 3 \\
		Learning strategy: 2\\
		Mito agglomeration: Off~~~~~~ \\  
		Threshold: 0.0\\~\\
	\end{tabular}
	&
	\begin{tabular}{l}
		\vspace{0.2mm} \\
		\midrule
		\emph{Results---Test Set} \\ \midrule Cost [m]: 217 \\ Val. acc.: 0.99 \\ Test. acc.: 0.68 \\ Prec./Recall: 0.58/0.56 \\ F1 Score: 0.54 \\
	\end{tabular}
\end{tabular}
\hrule
}
\label{tab:parameters}
\end{table}

\subsection{Baseline Comparisons}

\paragraph{Interactive proofreading.} Haehn~\etal's comparison of interactive proofreading tools concludes that novices perform best when using Dojo~\cite{haehn_dojo_2014}. We studied the publicly available findings of their user study and use the data of all Dojo users in aggregate as a baseline.

\paragraph{Computer-aided proofreading.} We compare against focused proofreading by Plaza~\cite{focused_proofreading}. Focused proofreading performs graph analysis on the output from NeuroProof~\cite{neuroproof2013}, instead of our GALA approach. Therefore, for training our focused proofreading baseline, we replace GALA in our automatic segmentation pipeline with NeuroProof but use exactly the same input data including membrane probabilities. We obtained the best possible parameters for NeuroProof by consulting the developers (Tab.~\ref{tab:parameters}). Rather than using Raveler as the frontend, we use our own interface (Fig.~\ref{fig:ui}) to compare only the classifier from Plaza's approach.

\subsection{Experiments}

\paragraph{Selection oracle evaluation.} We use the selection oracle as described in Sec.~\ref{sec:errorcorrection} for the decision whether to accept or reject a correction. The purpose of this experiment is to investigate how many corrections are required to reach the best possible outcome. This is a direct comparison of the guided proofreading and focused proofreading classifiers but can only be performed if ground truth data is available. We perform this experiment on all datasets listed above.

\paragraph{Automatic method evaluation.} For this experiment, we accept all suggested corrections if the rankings are above a configured threshold $p_t=.95$ (Sec.~\ref{sec:errorcorrection}). We observed this value as stable in previous experiments with the guided proofreading classifiers (see supplementary material). We compare against the focused proofreading classifier and perform this experiment on all reported datasets.

\paragraph{Forced choice user experiments.} We conducted a quantitative user study to evaluate the forced choice setting (Sec.~\ref{sec:errorcorrection}). In particular, we evaluated how participants perform while correcting an automatic segmentation using the guided proofreading and focused proofreading tools. We designed a single factor between-subjects experiment with the factor \textit{proofreading classifier}, and asked participants to proofread the AC4 subvolume in a fixed time frame of 30 minutes. To enable comparison against the interactive proofreading study by~Haehn~\etal~\cite{haehn_dojo_2014}, we use the exact same study conditions, dataset, and time limit. The experiment was performed on a machine with standard off-the-shelf hardware. All participants received monetary compensation.

\paragraph{Novice study design.} We recruited participants with no experience in electron microscopy data or proofreading through flyers, mailing lists, and personal interaction. Based on sample size calculation theory, we estimated the study needed ten users per proofreading tool including four potential dropouts~\cite{samplesize1,samplesize2}. All twenty participants completed the study ($N=20$, 10 female; 19-65 years old, $M$=30). 

Each study session began with a five minute standardized explanation of the task. Then, the participants were asked to perform a 3 minute proofreading task on separate but representative data using focused proofreading. The participants were allowed to ask questions during this time. The classifier did not matter in this case since the user interface was the same. The experimenter then loaded the AC4 subvolume with initial pre-computed classifications by either guided proofreading or focused proofreading depending on assignment. After 30 minutes, the participants completed the raw NASA-TLX standard questions for task evaluation~\cite{NASATLX}.
\begin{figure*}[t]
\centering
\includegraphics[width=\linewidth]{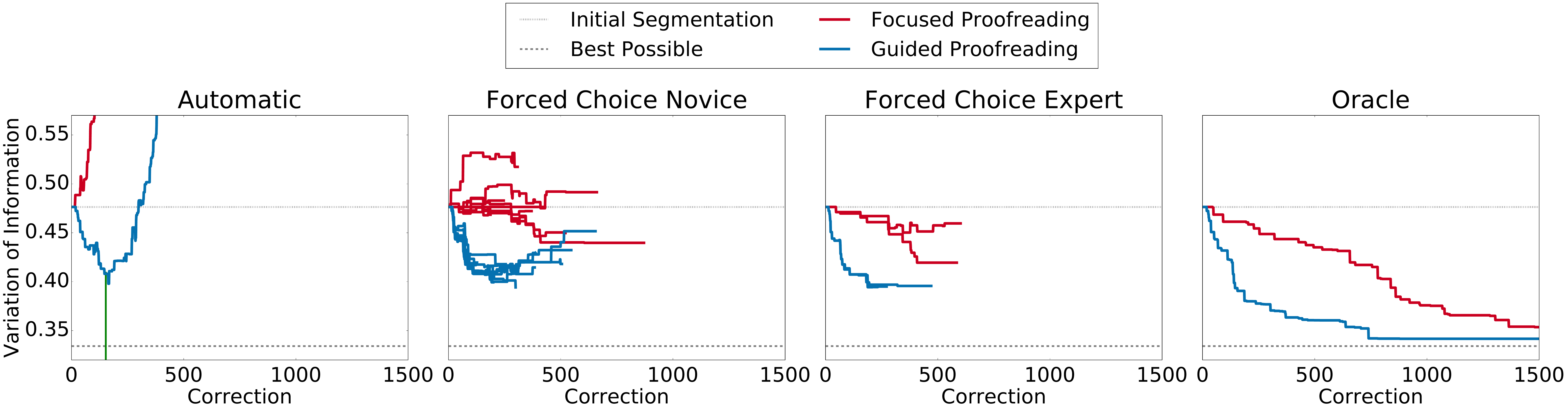}
\caption{Performance comparison of Plaza's focused proofreading (red) and our guided proofreading (blue) on the AC4 subvolume. All measurements are reported as median VI, the lower the better. We compare different approaches of accepting or rejecting corrections for each method: automatic selection with threshold (green line), forced choice by ten novice users, forced choice by two domain experts, and the selection oracle. In all cases, guided proofreading yields better results with fewer corrections.}
\label{fig:ac4trails}
\end{figure*}

\paragraph{Expert study design.} We recruited 4 domain experts to evaluate the performance of both guided and focused proofreading. We obtained study consent and randomly assigned 2 experts to proofread using each classifier. The experts performed the 3 minute test run on different data prior to proofreading for 30 minutes. After the task ended, the experts were asked to complete the raw NASA-TLX questionnaire.

\paragraph{Evaluation metric.} We measure the similarity between proofread segmentations and the manual `ground truth' labelings using \textit{variation of information} (VI). VI is a measure of the distance between two clusterings, closely related to mutual information (the lower, the better).

\section{Results and Discussion}

Additional plots are available as supplemental material due to limited space. 

\subsection{Classification Performance}

\begin{figure}[t]
\centering
\includegraphics[width=0.9\linewidth]{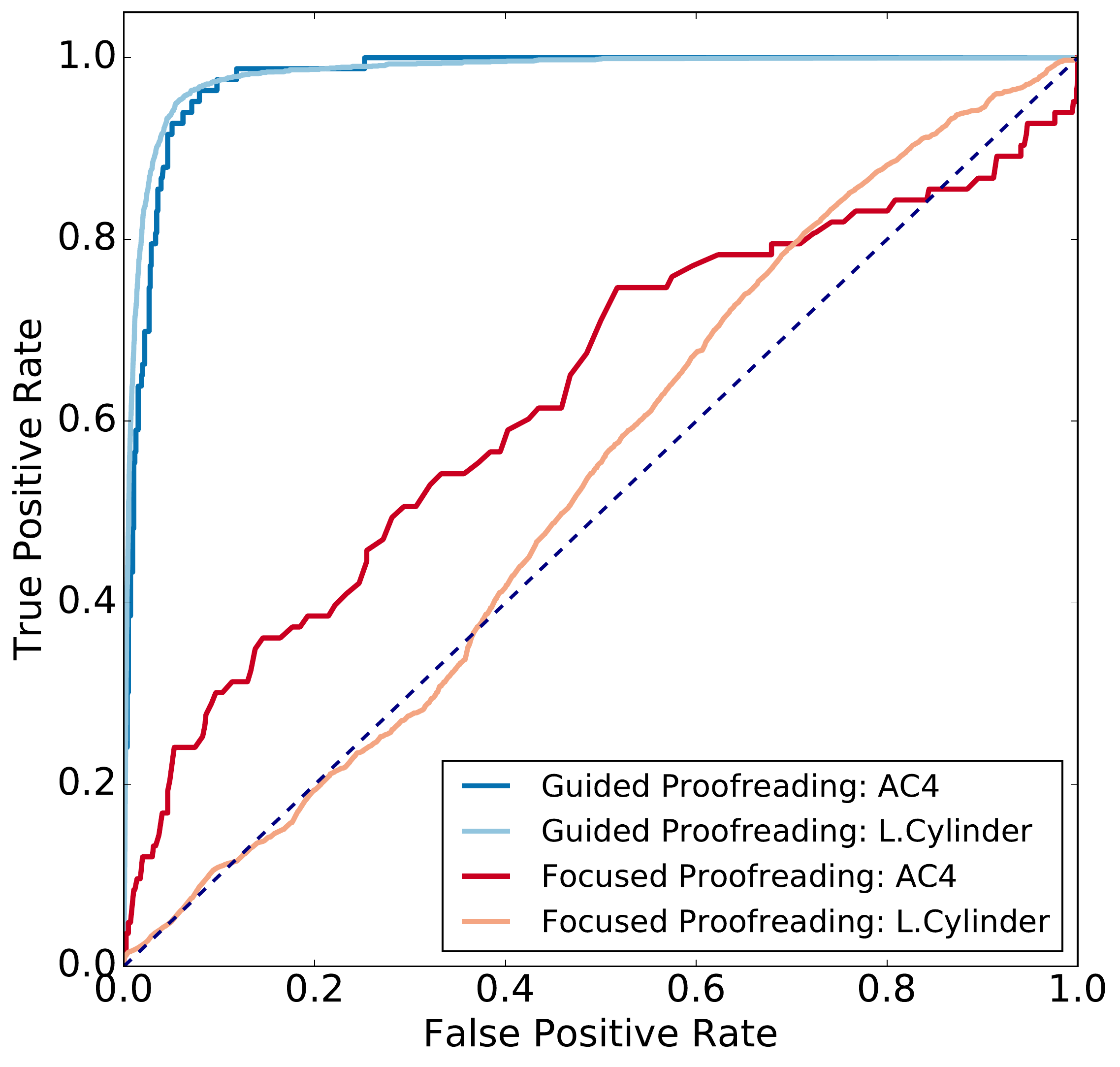}
\caption{Receiver Operating Characteristic curves comparing focused proofreading and guided proofreading automatic correction. We evaluate on unbalanced test sets of the AC4 subvolume (darker colors) and the L. Cylinder volume (lighter colors). Guided proofreading performs better.}
\label{fig:pr}
\end{figure}

\paragraph{L.~Cylinder.} Evaluation was performed on previously unseen sections of the mouse cortex volume from Kasthuri~\etal~\cite{kasthuri2015saturated}. We generated a dataset of 81,184 correct and 8,780 split error patches with respect to the ground truth labeling. Then, we classified each patch by using focused proofreading and guided proofreading, and compare performance (Fig. \ref{fig:pr}). Our method exhibits higher sensitivity and lower fall-out.

\paragraph{AC4 subvolume.} We generated 3,488 correct and 332 error patches (10 merge errors, 322 split errors). Guided proofreading achieves better classification performance (Fig. \ref{fig:pr}).

\subsection{Forced Choice User Experiment}
We performed a user study to evaluate the forced choice error correction method among novices and experts. To be comparable to Haehn~\etal's Dojo user study~\cite{haehn_dojo_2014}, participants were asked to proofread the AC4 subvolume for 30 minutes. We counted 10 merge errors and 322 split errors by computing the maximum overlap of the initial segmentation with respect to the ground truth labeling (provided in \cite{haehn_dojo_2014}). For evaluation, we measure the performance of proofreading quantitatively by comparing VI scores of segmentations. Median VI $=0.476$ ($SD=0.089$), with mean VI $=0.512$ ($SD=0.09$). Most novices and all experts were able to improve upon this score with both focused proofreading and guided proofreading (Fig.~\ref{fig:ac4trails}).

\paragraph{Novice performance.} Participants using focused proofreading were able to reduce the median VI of the automatic segmentation to $0.469$ ($SD=0.87$). On average, users viewed $423.4$ corrections and accepted $45.8$.
Participants using guided proofreading were able to reduce the median VI to $0.424$ ($SD=0.037$). Here, users viewed on average $353.4$ corrections and accepted $106.9$.
While three users of focused proofreading made the initial segmentation worse, all participants using guided proofreading were able to improve it. In comparison to the results of Haehn~\etal, focused and guided proofreading outperform interactive proofreading with Dojo (median VI $0.535$, $SD=0.055$). The slope of VI score per correction (Fig.~\ref{fig:ac4trails}) and average timings (Tab.~\ref{tab:correctiontimes}) show that guided proofreading enables improvements with fewer corrections than the other tools. Interestingly, novice performance decreases after approximately $300$ corrections. There are two explanations for this: user fatigue, and increasing uncertainty during error suggestion from the classifier. 

\begin{table}[t]
\caption{Average proofreading speed for novice users of Dojo, Focused Proofreading (FP) and our Guided Proofreading (GP). Our system achieves significantly higher VI reduction per minute (7.5$\times$) over state-of-the-art FP, while being slightly slower per correction.}
\resizebox{\linewidth}{!}{
\begin{tabular}{lrrrr}
\toprule
\makecell{Approach\\(Novice)} & \makecell{Time Per\\Correction (s)} & \makecell{VI Reduction\\Per Minute} & \makecell{Improvement} \\
\midrule
\emph{Dojo} & 30.5 & -0.00200 & $-8.7\times$ \\
\emph{FP} & 4.9 & 0.00023 & $1.0\times$\\
\emph{GP} & 6.2 & 0.00173 & $7.5\times$\\
\bottomrule
\end{tabular} 
}
\label{tab:correctiontimes}
\end{table}

\paragraph{Expert performance.} Domain experts were able to improve the initial segmentation. With focused proofreading, the median VI of the automatic segmentation was $0.439$ ($SD=0.084$). With guided proofreading, the median VI was $0.396$ ($SD=0.032$, Fig.~\ref{fig:ac4boxplot}).

\paragraph{Subjective responses.} We used the NASA-TLX workload index to record subjective responses. Mental, physical, and temporal demands were reported slightly higher for participants using focused proofreading. However, these differences were not statistically significant. This is unsurprising as the user interface was the same for both groups.

\subsection{Automatic Error Correction}

\begin{figure}[t]
\centering
\includegraphics[width=\linewidth]{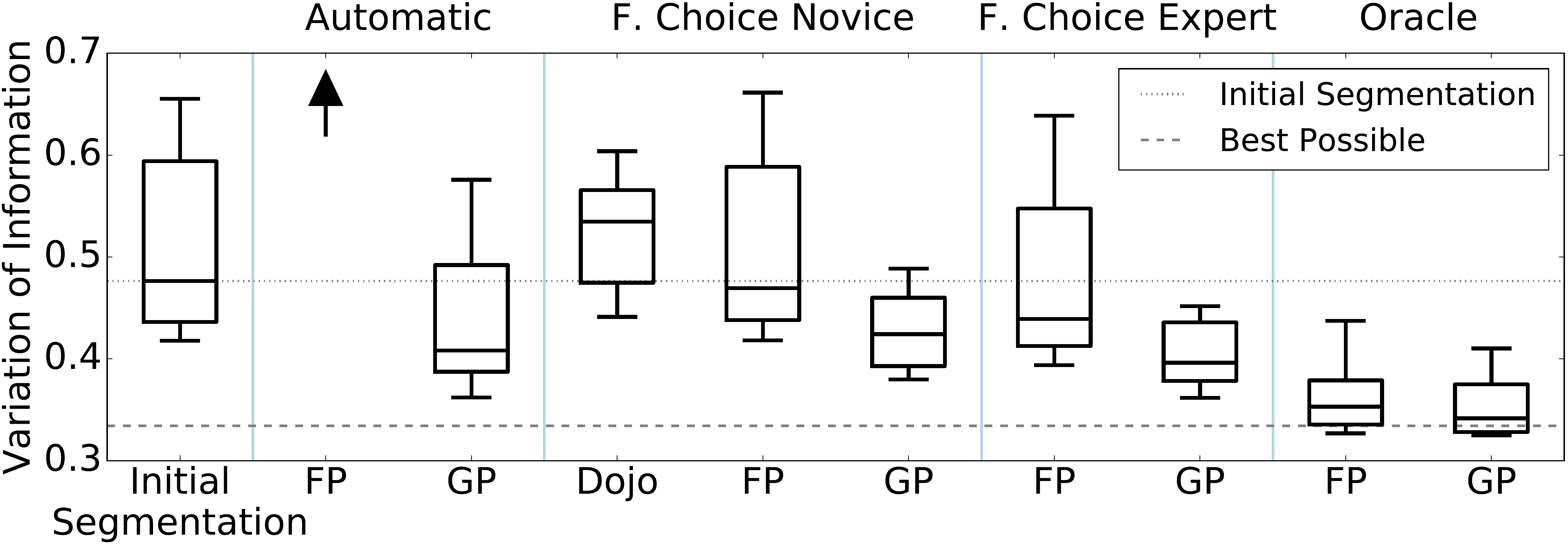}
\caption{VI distributions of guided proofreading (GP), focused proofreading (FP) and Dojo output across slices of the AC4 subvolume, with different error correction approaches. The variation resulting from performance of FP with automatic selection is $4.5\times$ higher than GP (as indicated by the arrow), with median VI of $1.9$ and $SD=0.496$.}
\label{fig:ac4boxplot}
\end{figure}

\paragraph{Selection oracle.} As expected, the selection oracle yields the best performance on all datasets. Fig.~\ref{fig:ac4trails} shows VI reduction using the selection oracle on the AC4 subvolume (initial median VI $0.476$, $SD=0.089$). With focused proofreading, the selection oracle reaches a median VI of $0.353$ ($SD=0.037$) after $1600$ corrections. With guided proofreading, the oracle reaches a minimum median VI of $0.342$ ($SD=0.03$) after $800$ corrections. Both results are close to the best possible median VI of $0.334$ (calculated by computing maximum overlap with the ground truth). The slope of the trails in Fig.~\ref{fig:ac4trails} shows that guided proofreading requires fewer corrections to reach a reasonable reduction in VI. Fig.~\ref{fig:ac4boxplot} shows the VI distribution across methods. On the L.~Cylinder dataset (initial VI $0.379$, $SD=0.118$), focused proofreading reduces the median VI to $0.298$ ($SD=0.075$) after $26,170$ corrections ($2,419$ accepted). Guided proofreading reaches the minimum median VI $0.2996$ ($SD=0.073$) after $10,000$ corrections (in total $27,491$, $2,696$ accepted).

\paragraph{Automatic selection with threshold.} Focused proofreading was not designed to run automatically. This explains the poor performance on the AC4 subvolume (VI of $1.9$, $SD=0.496$) and on the L.~Cylinder dataset (VI of $2.75$, $SD=0.789$). For guided proofreading, we set $p_t=0.95$ for both datasets. This reduces median VI in the AC4 subvolume to $0.398$ ($SD=0.068$). This result is comparable to expert performance. Guided proofreading also reduces VI in the L.~Cylinder data to $0.352$ ($SD=0.087$).

\paragraph{Merge Error Detection.} Guided proofreading performs merge error detection prior to split error detection. The classifier found 10 merge errors in the AC4 subvolume, of which 4 reduced VI. Automatic selection with $p_t=0.95$ corrected 6 of these errors (Prec./Recall 0.87/0.80, F1-score 0.80). This was not captured in median VI, but resulted in a mean VI reduction from $0.512$ ($SD=0.09$) to $0.509$ ($SD=0.086$). The selection oracle reduced mean VI with only merge errors to $0.508$ ($SD=0.086$). In the forced choice user study, novices marked 1.9 merge errors for correction and reduced mean VI to $0.502$ (experts marked 2, VI $0.503$, $SD=0.086$). This shows how hard it is to identify merge errors. In 50 sections of the L.~Cylinder dataset, 151 merge errors were automatically found of which 17 reduced VI. Automatic selection with $p_t=0.95$ corrected 6 true VI-reducing errors and 30 VI-increasing ones (Prec./Recall 0.82/0.73, F1-score 0.77) to negligible VI effect.

\section{Conclusions}

Humans are the bottleneck when proofreading segmentation data and minimizing the manual labor is the goal. Our classifiers suggest potential errors and corrections better than existing methods. This reduces the time spent finding and correcting errors.
Our experiments also show that automatic proofreading has potential to further reduce human involvement. This will be the target of future research. We provide our framework and data as free and open research at \url{http://rhoana.org/guidedproofreading/}.

\section*{Acknowledgements}
We would like to thank Stephen Plaza for detailed explanations of focused proofreading and Toufiq Parag for the configuration of the NeuroProof classifier.

{\small
\bibliographystyle{ieee}
\bibliography{connectomics}
}

\end{document}